\documentclass[
]{ceurart}

\usepackage{listings}
\usepackage{algorithm}%
\usepackage{algorithmicx}%
\usepackage{algpseudocode}%
\begin{document}

\copyrightyear{2025}
\copyrightclause{Copyright for this paper by its authors.
  Use permitted under Creative Commons License Attribution 4.0
  International (CC BY 4.0).}

\conference{RuleML+RR'25: Companion Proceedings of the 9th International Joint Conference on Rules and Reasoning, September 22--24, 2025, Istanbul, Turkiye}

\title{Investigating Language Model Capabilities to Represent and Process Formal Knowledge: A Preliminary Study to Assist Ontology Engineering}

\author[1]{Hanna Abi Akl}[%
orcid=0000-0001-9829-7401,
email=hanna.abi-akl@inria.fr,
url=https://hannaabiakl.github.io/,
]
\address[1]{Supervised by Fabien Gandon, Catherine Faron and Pierre Monnin\\
Université Côte d’Azur, Inria, CNRS, I3S, Sophia Antipolis, France and Data ScienceTech Institute (DSTI), Paris, France}


\begin{abstract}
  Recent advances in Language Models (LMs) have failed to mask their shortcomings particularly in the domain of reasoning. This limitation impacts several tasks, most notably those involving ontology engineering. As part of a PhD research, we investigate the consequences of incorporating formal methods on the performance of Small Language Models (SLMs) on reasoning tasks. Specifically, we aim to orient our work toward using SLMs to bootstrap ontology construction and set up a series of preliminary experiments to determine the impact of expressing logical problems with different grammars on the performance of SLMs on a predefined reasoning task. Our findings show that it is possible to substitute Natural Language (NL) with a more compact logical language while maintaining a strong performance on reasoning tasks and hope to use these results to further refine the role of SLMs in ontology engineering.
\end{abstract}


\maketitle

\section{Introduction}
\label{section-introduction}


Since their introduction, Language Models (LMs) have achieved impressive results on a broad spectrum of Natural Language Processing (NLP) tasks, ranging from text classification to question-answering and text generation \cite{guo2023evaluating, minaee2402large}. One field that still poses a challenge for LMs is reasoning. Over different kinds of reasoning tasks, LMs still struggle to achieve comparable results to humans in tasks involving explicit or implicit thinking (e.g. mathematical, commonsense, logical first-order reasoning) \cite{guo2023evaluating, liu2025logical}. These reasoning capabilities are essential for complex tasks such as knowledge graph construction or ontology engineering \cite{wang2024towards, jin2024large}. For the latter, LMs have been used in various roles in the ontology learning pipeline, from requirement gathering to ontology creation, validation and alignment \cite{lippolis2025ontology, zhao2024improving, zhao2024using, zhao2025leveraging, fathallah2024neon}. Current research promotes neuro-symbolic approaches integrating symbolic mechanisms to curb LMs as current fine-tuning techniques (e.g. prompting) have proven insufficient in making them reliable reasoners \cite{hou2025neural, schlegel2022can}.

In this perspective, the goal of our research in this PhD is to evaluate reasoning in LMs by focusing on small language models (SLMs) for the frugality, autonomy and privacy they support and by narrowing our evaluation to first-order logic (FOL) reasoning. We present and evaluate an experimental setup for testing the performance of SLMs on classifying the satisfiability of a logical conclusion given a set of premises. We also experiment with different knowledge representation languages to measure the impact of the formal representation of the input on the performance of SLMs on reasoning tasks. In doing so, we aim to address the following research question (RQ): \textit{RQ: Is there a better formal representation for logical data than natural language?}

This paper is structured as follows: Section \ref{section-stateofart} provides a review of the state of the art in LM reasoning methods in the context of formal problem resolution with a focus on logic tasks. Section \ref{section-methodology} introduces our methodology and formalizes the Common Logic Grammar Construction (CLGC) pipeline. Section \ref{section-experiments} presents our extended experimental setup and reports our findings with discussions on our results. Finally, Section \ref{section-conclusion} provides a summary of our key findings and outlines the next steps in this PhD research as well as potential future research directions.

\section{State of the Art}
\label{section-stateofart}

The focus of the first stage of this PhD is to combine formal knowledge representation with LM learning methods to improve LM performance on FOL reasoning tasks with the perspective of an application to ontology engineering. We report here the relevant research regarding state-of-the-art LM reasoning methods and logical reasoning applications.

\subsection{Language Model Reasoning}
\label{subsection-lmreasoning}

Reasoning is a fundamental aspect of human intelligence and is a cognitive process that involves the use of evidence, arguments and logic to arrive at conclusions \cite{huang2022towards}. The impressive performance of LMs on Natural Language (NL) data has led to observations that these models may exhibit reasoning abilities when they are sufficiently large \cite{huang2022towards}. However, reasoning also appears to be inherent in SLMs which are shown to be able to perform close to, and in some cases on par with, Large Language Models (LLMs). Approaches include extensive pre-training or guided techniques such as problem decomposition or the introduction of LLMs in SLM training processes for step correction as in the case of the SMART framework \cite{srivastava2025towards, bi2024enhancing, kim2025guiding}.

In an effort to determine reasoning abilities in LMs, researchers have explored novelties in techniques, benchmark datasets and evaluation metrics \cite{huang2022towards}. Reasoning techniques largely fall under two pillars: LM-based and Reinforcement Learning (RL) techniques \cite{besta2025reasoning}. LM-based reasoning techniques revolve around the methodologies spearheading model training, i.e. Supervised Fine-Tuning (SFT) on the one hand and Prompting and In-Context Learning on the other \cite{huang2022towards, kumar2025llm}. Another emerging research direction is the exploration of hybrid or neuro-symbolic models which incorporate a form of symbolic knowledge into LMs to guide or enhance reasoning as seen in DomiKnowS and Logic \cite{faghihi2021domiknows, kesseli2025logic}.
Different training techniques extending fine-tuning and prompting like COCONUT have resulted in LMs being used to tackle various reasoning tasks like mathematical, logical, causal, commonsense reasoning and coding \cite{hao2022training}. None of these methods however consider the potential impact of the logical formalism used to prompt the LMs.


\subsection{Logical Reasoning Tasks for LMs}
\label{subsection-foltasks}

Logical reasoning is an interesting domain of application for LMs since it subjects them to different types of reasoning, namely deductive, inductive, abductive or analogical \cite{huang2022towards}. Reasoning problems are classified in the literature into 2 main groups: Logical Question Answering (LQA) and Logical Consistency (LC) \cite{cheng2025empowering}. LQA applications encompass tasks that require LMs to generate the correct answer within complex logical problems which require sophisticated reasoning given a collection of premises and constraints \cite{cheng2025empowering}. LC applications require LMs to generate verifiable answers to complex questions without violating consistency (e.g. the model should not generate answers that contain contradictions) \cite{cheng2025empowering}. As we will show, our first research questions, methods and experimental setup fall under LQA and rely on FOLIO, a human-annotated benchmark dataset for FOL reasoning \cite{han2024folio}.

LQA models can be grouped into 3 categories depending on the approach. Solver-aided models rely on LMs translating NL problems into FOL before passing them to a solver to compute the final answer. These models have shown an increased performance over traditional model training techniques on logical problem-solving \cite{zhang2022evaluating, xu2024faithful}. Prompt-based models rely on techniques like Chain-of-Thought (CoT) to break down complex problems using smaller reasoning steps (i.e. hops) and guide the model to the final solution \cite{yang2024large}. However, experiments on these models did not prove actual reasoning beyond two hops and could not determine that this type of reasoning scales with larger models \cite{yang2024large, yang2024larger}. The last group of models adhere to the pre-training and fine-tuning approaches on logical questions to enhance their reasoning process. These models often include an additional neural layer that captures logical constraints and benefits from forward and backward passes to iteratively refine the model output in the constrained setting \cite{ghosh2024logical}. Our methodology targets the data representation of the input and evaluates models with different representations in both the prompting and fine-tuning settings. Research on input representation has already shown that the language in which a LM receives a problem has an impact on its success rate in solving the problem \cite{peng2024playing}, but does not consider the variety of languages we evaluate in our work.


\section{SEF-CLGC Methodology}
\label{section-methodology}

This section introduces our methodology in three steps: (1) we characterize and categorize the different syllogisms that we will confront the LMs with, then (2) we introduce the different grammars we will use and evaluate in solving the logical reasoning tasks and (3) we provide an overview of the pipeline implementing the complete experiment. The following sub-sections present the Syllogistic Evaluation Framework (SEF) and Common Logic Grammar Construction (CLGC) pipeline which are integral components for studying the impact of different formal representations on reasoning tasks as part of this PhD research. The last sub-section illustrates the process to derive the CLGC pipeline as well as the generated grammars which constitute the crux of our experimental studies.

\subsection{Syllogistic Evaluation Framework to Characterize the Logical Reasoning Tasks}
\label{subsection-sef}

The Syllogistic Evaluation Framework (SEF) is a methodology for identifying and classifying different types of syllogisms inspired by the work in \cite{wu2023hence}. The objective of SEF is to have an additional evaluation criterion to trace and analyze performance on different types of logical reasoning: SEF allows discrimination between different reasoning problems to explicit a model's reasoning capabilities.
We applied SEF on the FOLIO dataset which is the focal data source for our experiments. FOLIO is a collection of stories, with each story $S$ consisting of a set of $n$ premises $P=\{p_1,p_2,...,p_n\}$ and $m$ conclusions $H=\{h_1,h_2,...,h_m\}$ in NL and their corresponding FOL annotations of $n$ premises $PF=\{pf_1,pf_2,...,pf_n\}$ and $m$ conclusions $HF=\{hf_1,hf_2,...,hf_m\}$. Given each premises-conclusions pair, the goal is to determine the truth values of the conclusions, i.e. ``True", ``False" or ``Uncertain" based on FOL reasoning.
We determined 4 syllogistic categories to classify each premises-conclusions pair: Disjunctive, Hypothetical, Categorical and Complex. The categories were identified based on the following logical criteria presented from highest to lowest order of precedence: (1) Disjunctive: Any premises-conclusions pair containing a disjunction $\vee$ or $\oplus$ (2) Hypothetical: Any premises-conclusions pair containing an implication $\implies$ (3) Categorical: Any premises-conclusions pair consisting of exactly 2 premises and a conclusion and (4) Complex: The default category for premises-conclusions pairs that do not fit in any of the three other categories.


\begin{table*}
  \caption{SEF classification examples on FOLIO.}
  \label{tab:example-syllogisms}
  \resizebox{\linewidth}{!}{
  \begin{tabular}{p{0.2\linewidth} |p{0.15\linewidth} | p{0.7\linewidth} | p{0.15\linewidth}|p{0.15\linewidth}}
    \toprule
    NL Premise&NL Conclusion&FOL Premise&FOL Conclusion&SEF Class\\
    \midrule
All squares are four-sided.
All four-sided things are shapes.&All squares are shapes.& \makecell{$\forall$x (Square(x) → FourSided(x)) \\ $\forall$x (FourSided(x) → Shape(x))}  
& $\forall$x (Square(x) → Shape(x))&Hypothetical\\
\hline
Some affection is love.
Some love is positive.&Some affection is positive.& \makecell{ $\exists$x (Affection(x) $\land$ Love(x)) \\  $\exists$x (Love(x) $\land$ Positive(x))}  
&  $\exists$x (Affection(x) $\land$ Positive(x))&Categorical\\
\hline
Diamond Mine is a professional wrestling stable formed in WWE.
Roderick Strong leads Diamond Mine.
Diamond Mine includes the Creed Brothers and Ivy Nile.
Imperium has a feud with Diamond Mine.&Roderick Strong leads the Creed Brothers.& \makecell{ProfessionalWrestlingStable(diamondMine) $\land$ In(diamondMine, WWE)\\
Leads(roderickStrong, diamondMine)\\
Includes(diamondMine, creedBrothers) $\land$ Includes(diamondMine, ivyNile)\\
Feuds(imperium, diamondMine)}&Leads(roderickstrong, creedbrothers)&Complex\\
\hline
Susan flies to LGA airport.
The departure and arrival can not be at the same airport.
John flies from LGA airport.&Susan flies from LGA airport.& \makecell{FlyTo(susan, lgaAirport)\\
$\forall$x $\forall$y (FlyFrom(x, y) $\oplus$ FlyTo(x, y))\\
FlyFrom(john, lgaAirport)} &FlyFrom(susan, lgaAirport)&Disjunctive\\
  \bottomrule
\end{tabular}
}
\end{table*}

\begin{table*}
  \caption{Syllogistic Evaluation Framework Statistics on FOLIO.}
  \label{tab:syllogismstats}
  \begin{tabular}{cccc}
    \toprule
    Syllogism&Train&Test&Valid\\
    \midrule
    Disjunctive & 414 & 113 & 99\\
    Hypothetical & 285 & 68 & 84\\
    Complex & 114 & 22 & 16\\
    Categorical & 6 & 2 & 4\\
  \bottomrule
\end{tabular}
\end{table*}


Examples of SEF classification are presented in Table \ref{tab:example-syllogisms}. The classification statistics for FOLIO are presented in Table \ref{tab:syllogismstats}. The distribution shows an imbalance heavily favoring Disjunctive and Hypothetical kinds of reasoning, while Complex and Categorical examples are too sparsely represented to draw any conclusions on.

\subsection{Alternative Languages for Transcribing the Logical Reasoning Tasks}
\label{subsection-clgcgrammars}

Different knowledge representation formalisms have been tested in the CLGC pipeline to measure the impact of the language on the SLM in solving a logical reasoning task. We present concrete transformations from FOL to CLGC languages in Table \ref{tab:example-transformations}.

\subsubsection{Common Logic}
\label{grammars-cl}

We implemented grammars from the Common Logic (CL) family suite as defined in \cite{sowa2008conceptual, sowa1992conceptual, sowa2011introduction}. In particular, we focused on the Common Logic Interchange Format (CLIF) and the Conceptual Graph Interchange Format (CGIF) languages for their ease of interoperability with FOL while retaining the advantages of formal logic representations. From a practical aspect, these languages can also be implemented due to the availability of their grammars\footnote{https://www.w3.org/DesignIssues/Sowa/cgstand.htm}. We limited our implementation to the core Backus–Naur form (BNF) grammars for CLIF and CGIF and did not include the extended syntax for these languages for simplicity and compactness. 


\subsubsection{Tensor Function Logic and Tensor Function Logic Plus}
\label{grammars-tfl}

We also incorporated Tensor Function Logic (TFL), a formal reasoning language introduced in \cite{sommers2017invitation, castro2018programming, manzano2019intermediate}, and its extension, Tensor Function Logic Plus (TFL+). We limited the scope of TFL to the terms and their arithmetic signs and incorporated quantifier subscript representation as well as parentheses in TFL+ to work with grammars of varying complexity. 

\subsubsection{MINIFOL}
\label{grammars-minifol}

Additionally, we implemented a custom language called Miniature First-Order Logic (MINIFOL) which is directly derived from FOL by replacing $\forall$ and $\exists$ with “all” and “some” respectively and replacing operators with Boolean equivalents (e.g. “$\wedge$” with “\&”). The advantage of MINIFOL is twofold: replace some FOL vocabulary (i.e. symbols) with vocabulary that is more familiar for SLMs (i.e. words) while benefiting from a rule set that enables an easy transformation from FOL to MINIFOL.

\begin{table*}
  \caption{Example transformations from FOL to CLGC languages.}
  \label{tab:example-transformations}
  \resizebox{\linewidth}{!}{
  \begin{tabular}{p{0.2\linewidth} | p{0.2\linewidth}|p{0.2\linewidth}|p{0.2\linewidth}|p{0.1\linewidth}|p{0.1\linewidth}}
    \toprule
    FOL&MINIFOL&CLIF&CGIF&TFL&TFL+\\
    \midrule
$\forall$x ((Employee(x) $\land$ Schedule(x, meeting, customers)) → AppearIn(x, company))&all:x ((employee(x) \& schedule(x, meeting, customers)) :- appearin(x, company))&forall x ((employee(x) and schedule(x, meeting, customers)) implies appearin(x, company))&[@every *x [([(employee[(?x)]  schedule[(?x  meeting  customers)])]  appearin[(?x  company)])]&-+E1++S1-+A1&-((+E0++S0)-+A0)\\
\hline
$\forall$x ((Employee(x) $\land$ HasLunch(x, company)) → Schedule(x, meeting, customers))&all:x ((employee(x) \& haslunch(x, company)) :- schedule(x, meeting, customers))&forall x ((employee(x) and haslunch(x, company)) implies schedule(x, meeting, customers))&@every *x [([(employee[(?x)]  haslunch[(?x  company)])]  schedule[(?x  meeting  customers)])]&-+E1++H1-+S1&-((+E0++H0)-+S0)\\
\hline
$\forall$x (Employee(x) → (HasLunch(x, company) $\oplus$ HasLunch(x, home)))&all:x (employee(x) :- (haslunch(x, company) \^ haslunch(x, home)))&forall x (employee(x) implies (haslunch(x, company) xor haslunch(x, home)))&@every *x [(employee[(?x)]  [(haslunch[(?x  company)]  haslunch[(?x  home)])])]&-+E1-+H1-+H1&-(+E0-(+H0-+H0))\\
\hline
$\forall$x ((Employee(x) $\land$ HasLunch(x, home)) → Work(x, home))&all:x ((employee(x) \& haslunch(x, home)) :- work(x, home))&forall x ((employee(x) and haslunch(x, home)) implies work(x, home))&@every *x [([(employee[(?x)]  haslunch[(?x  home)])]  work[(?x  home)])]&-+E1++H1-+W1&-((+E0++H0)-+W0)\\
\hline
$\forall$x ((Employee(x) $\land$ (¬In(x, homecountry))) → Work(x, home))&all:x ((employee(x) \& (~in(x, homecountry))) :- work(x, home))&forall x ((employee(x) and (not in(x, homecountry))) implies work(x, home))&@every *x [([(employee[(?x)]  [(~in[(?x  homecountry)])])]  work[(?x  home)])]&-+E1+-+I1-+W1&-((+E0+(-+I0))-+W0)\\
\hline
$\forall$x (Manager(x) → ¬Work(x, home))&all:x (manager(x) :- ~work(x, home))&forall x (manager(x) implies not work(x, home))&@every *x [(manager[(?x)]  ~work[(?x  home)])]&-+M1--+W1&-(+M0--+W0)\\
\hline
¬(Manager(james) $\oplus$ AppearIn(james, company))&~(manager(james) $\land$ appearin(james, company))&not (manager(james) xor appearin(james, company))&~[(manager[(james)]  appearin[(james  company)])]]&-+M1-+A1&-(+M2(+j2)-+A2)\\
\hline
HasLunch(james, company)&haslunch(james, company)&haslunch(james, company)&[haslunch[(james  company)]]&+H1&+H2\\
  \bottomrule
\end{tabular}
}
\end{table*}

\subsection{CLGC Pipeline and Alternative Configurations}
\label{subsection-clgcpipeline}

\begin{figure}
  \centering
  \includegraphics[width=0.6\linewidth]{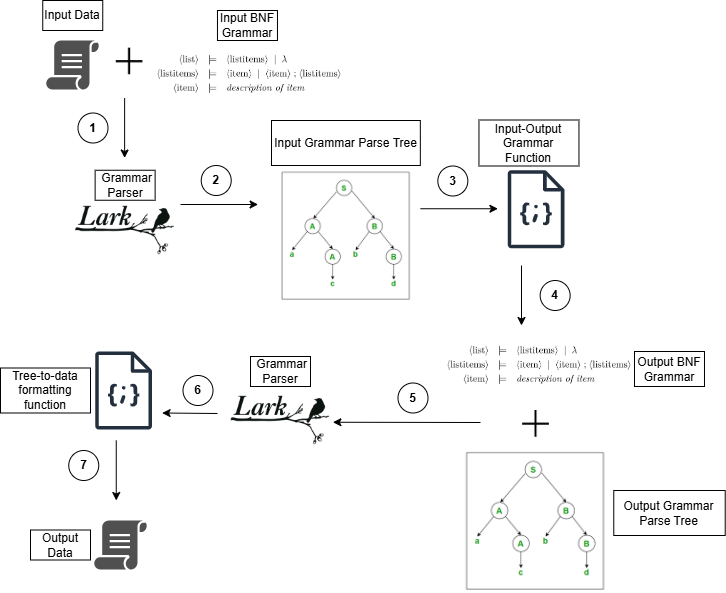}
  \caption{CLGC pipeline.}
  \label{fig:clgc-pipeline}
\end{figure}
\begin{figure}
  \centering
  \includegraphics[width=1.1\linewidth]{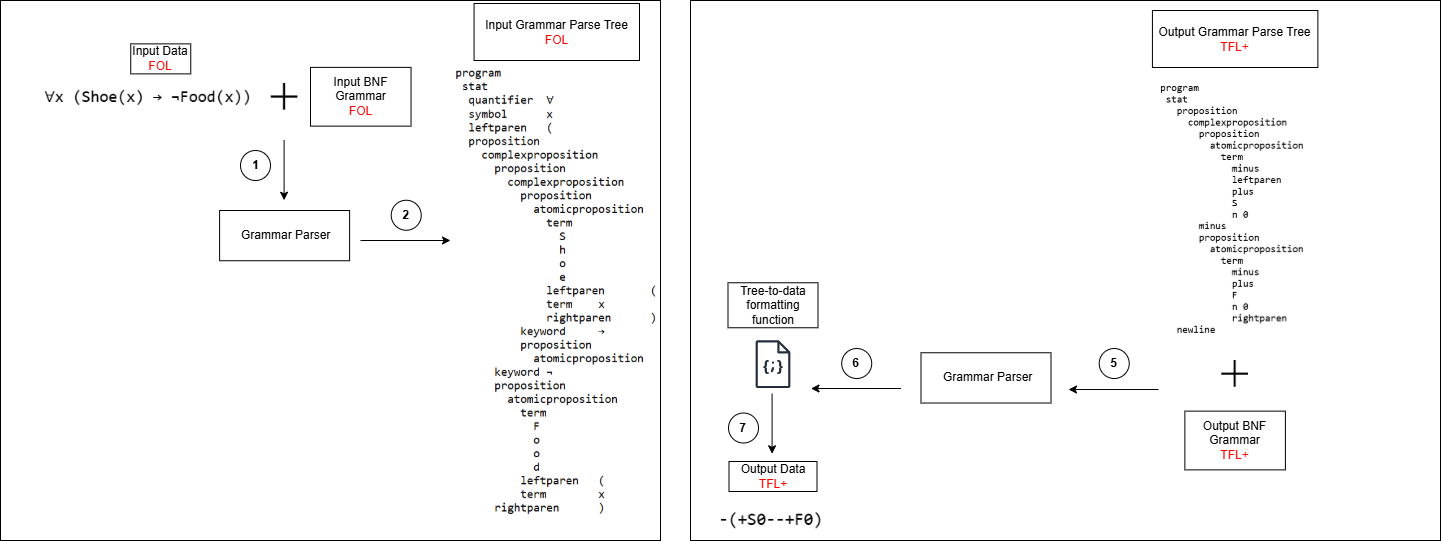}
  \caption{(Left) Input-to-tree example for Steps 1 and 2 of the CLGC pipeline. (Right) Tree-to-output example for Steps 5, 6 and 7 of the CLGC pipeline. The text in red represents the grammar of choice for the example.}
  \label{fig:clgcexample}
\end{figure}

Because this work investigates the impact of the formal representation of knowledge in the scope of reasoning tasks, we devised a pipeline to transform one logical form to another with the goal of generating different data representations from FOLIO to test as input to our SLMs. The CLGC pipeline is presented in Figure \ref{fig:clgc-pipeline} and consists of the following steps: (1) FOLIO input data (i.e. a premises-conclusions pair) is passed along with an input grammar (i.e. FOL) in BNF to a parser. We used the Python Lark\footnote{https://lark-parser.readthedocs.io/en/stable/} parser in our work. (2) The parser generates an input grammar parse tree for the input language. Figure \ref{fig:clgcexample} shows a concrete example of generating the corresponding tree for a FOL input. (3) The parse tree is passed to an input-output grammar function which generates a new parse tree in the desired language (e.g. CLIF). The algorithm for the transformation is detailed in Algorithm \ref{algo1}. The implementation relies heavily on the BNF grammar definition of the input and target languages as it uses the non-terminal symbol definitions of both grammars to construct the target parse tree from the input tree by identifying the corresponding mappings and augmenting the target tree recursively. (4) The algorithm either successfully generates the parse tree in the target language or returns an error. In the event of the latter, the input and target BNF grammars are reviewed and corrected manually. (5) The target parse tree is passed along with its corresponding language BNF grammar to the Lark parser. (6) The parser generates an output string corresponding to the output data (i.e. premises-conclusions pair) in the target language. The output is passed to a formatting function to align the syntax with the target language (e.g. spacing). (7) The pipeline outputs a correctly-formatted premises-conclusions pair in the target language. Figure \ref{fig:clgcexample} summarizes Steps 5-7.

\begin{algorithm}
\caption{Input-Output Grammar Transformation Algorithm}\label{algo1}
\begin{algorithmic}[1]
\State \Require $G_i := L_i$
\Statex $D_i \in \{L_i\}$
\Statex $D_o \in \{L_o\}$
\Statex $G_o := L_o$
\State \Ensure $|L_i| < \infty$
\Statex $|L_o| < \infty$
\Statex $D_i \neq \varnothing$
\Statex $D_o = \varnothing$ 
\For{$d \in D$} \algorithmiccomment{loop through logical statements}
\State $P_i \Leftarrow generateParseTree(d, G_i)$ \algorithmiccomment{generate grammar parse tree from statement}
\State $P_o \Leftarrow \varnothing$ \algorithmiccomment{initialize output grammar parse tree for statement}
\State \For{$p \in P_i$} \algorithmiccomment{loop through each element of the input grammar parse tree}
\State $o \Leftarrow mapToParseTree(p, G_o)$ \algorithmiccomment{map input element to corresponding output grammar element}
\If{$o \neq \varnothing$} \algorithmiccomment{there is a corresponding element in the grammar mapping}
            \State $P_o \Leftarrow o$ \algorithmiccomment{add mapped element to output grammar parse tree}
            \Else \algorithmiccomment{there is no corresponding element in the grammar mapping}
            \State $continue$
\EndIf
\EndFor
\State $v \Leftarrow validateParseTree(P_o, G_o)$ \algorithmiccomment{check if final parse tree from statement is parsable}
\If{$v = True$}
            \State $D_o \Leftarrow P_o$
        \Else
            \State $G_o \Leftarrow updateGrammar(P_o, G_o)$ \algorithmiccomment{update output grammar to parse mapped parse tree}
        \EndIf
\EndFor
\end{algorithmic}
\end{algorithm}

Additionally, we introduced 2 new SFT configurations: Grammar Context-Passing (ICGP) and Tokenizer Re-Training. In ICGP, the BNF grammar of the input data language is passed as additional context to the classical SFT input (i.e. the logical premises, conclusions and label). In the Tokenizer Re-Training setup, the SLM tokenizer is retrained on the vocabulary of the grammar by passing all train, test and validation data in the specified language to the model. After re-training, the original size of the tokenizer is kept in some runs to incorporate the vocabulary of the grammar and sufficiently resized empirically in others to eliminate as many tokens outside those making up the target vocabulary.
 
\section{Experiments and Results}
\label{section-experiments}

In this section, we describe our experimental setup and set out to answer the following research sub-questions (SRQs):  \textit{SRQ1: Which training method yields the best results for solving FOL problems with LMs?} \textit{SRQ2: How do formal representations scale with models?} \textit{SRQ3: Does having a more compact vocabulary boost model performance for formal languages?}

\subsection{Experimental Setup}
\label{subsection-experimentalsetup}

We extend the experiments performed in \cite{han2024folio} on the NL reasoning with FOL task by expanding the data representation of the premises and conclusions as well as the choice of models. For the data representation, we retain the NL and FOL languages in addition to the ones we derived from our pipeline in section \ref{section-methodology} (i.e. CLIF, CGIF, TFL, TFL+ and MINIFOL). For our model selection, we restricted our choices to SLMs for their frugality and promising performances on the original task in \cite{han2024folio} and because they avoid having a dependency to external and possibly expensive LLM services. We configure our models (\textit{M}), grammars (\textit{G}) and learning methods (\textit{L}) with the values: \textit{M} = \{Flan-T5-small, Flan-T5-base, Flan-T5-large, GPT-2, Phi-3.5-mini-instruct, Gemma-2-2b-it\}, \textit{G} = \{NL, FOL, CLIF, CGIF, TFL, TFL+, MINIFOL\} and \textit{L} = \{SFT, Zero-Shot (ZS) Prompting, Few-Shot (FS) Prompting\}.

Our experimental runs consisted of variations of the triple (\textit{M}, \textit{G}, \textit{L}) to determine the truth value of a conclusion (i.e. ``True", ``False" or ``Uncertain") based on FOL reasoning over a set of premises. We split the provided train portion of the original version of FOLIO\footnote{https://huggingface.co/datasets/yale-nlp/FOLIO} into 80\% train and 20\% test and froze those splits along with the original validation set for all our experimental runs. Our final splits consisted of 800 training, 203 validation and 201 test rows.

All experiments were run on Google Colab instances using L4 GPUs and, depending on availability, A100 High-RAM GPUs. In order to run some configurations in the absence of A100 GPUs (e.g. SFT with Flan-T5-large), we used Parameter-Efficient Fine-Tuning with Low-rank adaptation (PEFT-LoRA). PEFT-LoRA freezes the pre-trained model weights to reduce the number of trainable parameters and has shown great efficacy for LM training \cite{hu2022lora}. In our setup, PEFT-LoRA is configured as follows: \{r=16, lora\_alpha=32, target\_modules=[``q", ``v"], lora\_dropout=0.05, bias=none", task\_type=``SEQ\_2\_SEQ\_LM"\}. In the SFT mode, all models were trained on 5 epochs without PEFT-LoRA and 10 epochs with PEFT-LoRA. 

For the ZS and FS modes, we used 2 prompt templates. The first is a basic template explaining the task, the nature of the input and the expected output. The second is the basic template augmented with the BNF grammar rules of the language when applicable (i.e. for languages other than NL). We performed 8-shot prompting for all FS experiments. Table \ref{tab:prompt-examples} shows an example of a prepared basic prompt and augmented prompt template for the ZS and FS experiments. No ablation study was performed on the CLGC pipeline as all components are needed to generate a correct grammar for our experiments.

\begin{table*}
  \caption{Example templates of basic and BNF grammar prompts. The \{...\} are replaced with real data.}
  \label{tab:prompt-examples}
  \resizebox{\linewidth}{!}{
  \begin{tabular}{c | p{0.8\linewidth}}
    \toprule
    Prompt Type&Example\\
    \midrule
    Basic & You are given a set of premises and a conclusion. The premises start after a <PREMISES> tag and conclusion starts after a <CONCLUSION> tag. Classify each conclusion as "True", "False" or "Uncertain" depending on its satisfiability of the premises.
<PREMISES>{...}</PREMISES>
<CONCLUSION>{...}</CONCLUSION>\\
\hline
    BNF Grammar & You are an expert in logic, and you need to output the conclusion for the following logic problem. You are given grammar rules for the language of the problem in <GRAMMAR> tags, and you should conform to these rules to understand the premises in the <PREMISES> tags and conclusion in the <CONCLUSION> tags. Classify each conclusion as "True", "False" or "Uncertain" depending on its satisfiability of the premises. Present your answer only in <output> tags.
<GRAMMAR>{...}</GRAMMAR>
<PREMISES>{...}</PREMISES>
<CONCLUSION>{...}</CONCLUSION>\\
  \bottomrule
\end{tabular}
}
\end{table*}


\subsection{Results}
\label{subsection-results}
We present the most prominent results from our experimental runs in Tables \ref{tab:sftresultssmall}-\ref{tab:syllogism-evaluation}. Table \ref{tab:sftresultssmall} shows that formalizing the premises and conclusions in CLIF ties the performance of NL on Accuracy and ranks second-best on F1 score. The result shows that SLMs can perform well on first-order reasoning tasks with a more compact formalism than NL. Languages with more complex syntaxes like CGIF struggle to yield good performances, suggesting that it is more difficult for SLMs to reason over complex representations. It is also worth noting that the best performances occur with the smallest model, Flan-T5-small, which outperforms larger, fine-tuned models (e.g. Flan-T5-large), suggesting that added architectural complexity may hinder the learning process in this setting. The result is empirically supported by the different data formalisms in our runs.

Table \ref{tab:sftresultslarge} shows scaled results on bigger models. With the A100 GPUs, models like Flan-T5-base and Flan-T5-large can be trained in supervised fashion without PEFT-LoRA tuning. The results confirm our earlier findings that a compact language like CLIF can effectively model logical data and be competitive on a reasoning task by ranking second-best to NL, i.e. the current state of the art in \cite{han2024folio}.

Table \ref{tab:zsresults} shows the impact of languages in the ZS setting. The results show that the performance of compact languages like CLIF and TFL+ are extendable to other models, as shown with Gemma-2-2b-it which ranks first in Precision with TFL+ and yields competitive results with CLIF. Not all grammars are created equally however, as shown by MINIFOL and TFL which for very small models like Flan-T5-small cripple the learning completely resulting in a flat performance of 0.

The results of Table \ref{tab:zsresultsgrammar} cement CLIF as the best language choice in the ZS setting. With and without BNF grammar prompting, CLIF outperforms all other languages. In addition, grammar prompting seems to improve SLM performance in the ZS case over first-order reasoning as empirically shown for all our languages. These results support similar findings in \cite{wang2023grammar}.

In Table \ref{tab:fsresults} we present our results for the 8-shot setting. The performances are comparable to the ZS case in Table \ref{tab:zsresults} and cement CLIF as a more compact contender to the verbose NL representation of logical input. A familiar pattern seen in the ZS case re-emerges as Flan-T5-small is unable to learn with TFL. This suggests a limitation to model learning with respect to size as even for very small grammars like TFL, very small models struggle to learn the formalism well enough to solve a reasoning task.

Table \ref{tab:fsresultsgrammar} showcases the effects of augmenting the FS prompt with the language BNF grammars. The first takeaway is that augmenting the basic prompt template with the BNF grammar does not seem to impact the performance of a model in the FS case, unlike the ZS setting in Table \ref{tab:zsresultsgrammar}. As for performance, the language choice does not seem to affect the model in the ZS setting, suggesting a model in that case might be less sensitive to the input representation. Finally, comparing these results to those in Table \ref{tab:zsresultsgrammar} shows that the same model performs better in the ZS case than the FS case. One explanation could be that the number of FS examples is not sufficient for the chosen model on this particular task for learning. Another plausible hypothesis can be that the examples might be skewing the learning of the model by adding noise in the case of solving a reasoning task.

Table \ref{tab:sftresultsgrammar} shows the results of passing BNF grammars as context in addition to the standard input in the SFT setting. Empirical results confirm that including the BNF grammar in-context hinders learning and degrades model performance on the first-order reasoning task. Model size and complexity does not seem to impact performance as scaling models still yields better results without the inclusion of the grammar. Comparing results to those in Tables \ref{tab:sftresultssmall} and \ref{tab:sftresultslarge} also shows weaker performance with in-context information passing than traditional supervised training on inputs.

Table \ref{tab:sftresultsvocab} shows the impact of re-training a model tokenizer to adapt to the specific vocabulary of a grammar and re-using the tokenizer in the SFT setting. The results show promise in the case of smaller models for more compact data representations. This is illustrated by the performance of Flan-T5-small on TFL+ with tokenizer re-training and vocabulary resizing which outperforms Flan-T5-small on TFL+ without tokenizer re-training and resizing. This configuration even outperforms Flan-T5-small and Flan-T5-large with PEFT-LoRA on NL from Table \ref{tab:sftresultssmall}, both of which represent the baseline of SFT on lower resources (i.e. L4 GPUs) for our task. However, as shown when moving to bigger models like Flan-T5-base and Flan-T5-large on TFL+, this method is not scalable and breaks down when tokenizer re-training is done, resulting in worse performances than using the default tokenizer for the models. One possible explanation is that re-training the tokenizer for small models and sufficiently compact data representations may make the models learn the formalism and solve the task more efficiently but also causes overfitting. This risk suggests that this method may not be adapted to take full advantage of the compactness of data representations as it sacrifices generalization and weakens overall performance on the reasoning task.

Finally, Table \ref{tab:syllogism-evaluation} uses SEF to showcase the predictive performances of our best configuration (i.e. Flan-T5-large in SFT setting) on the 3 best grammars (i.e. NL, CLIF and TFL+) on the FOLIO validation set. In all 3 cases, the model performs well on the Disjunctive and Hypothetical syllogisms which can be seen as an unsurprising result given the over-representation of these two types in FOLIO. Irrespective of the data representation, the model seems to also perform well on the Complex syllogisms, with TFL+ even yielding a slightly better performance on that type than NL and CLIF. Complex syllogisms typically contain 3 or more premises and polysyllogisms which might render them too ambiguous for NL and CLIF and may benefit more compact representations like TFL+. This may explain the slight increase in performance noted in the results. As for the Categorical syllogisms which are heavily under-represented, all grammars share equal performance with 2 hits and 2 misses on the prediction labels, but the data imbalance makes it difficult to interpret the results. Further investigation is needed on the quality of these predictions, especially in a comparative framework, to assess the capabilities of the model on different data representations. We provide an example of extended error analysis in Table \ref{tab:syllogismerroranalysis}. The analysis is conducted for the Categorical syllogistic type and shows that the variation in grammar generally results in the same model reasoning on these problems with the exception of TFL+ on a single occurence, whereby the model reasons "False" as opposed to "Uncertain" on NL and CLIF. A form of quantitative analysis may be needed to explore these results further and may be included in future work along with the analysis of the other syllogistic types which has not been included due to size restriction. The results enable us to answer our SRQs. For SRQ1, the best method shown empirically remains SFT as it provides the highest and most stable results among different representations. For SRQ2, our results show that controlled formal languages generally scale well with models while remaining consistent with performances. For SRQ3, our experiments show that tailoring the vocabulary size of the model tokenizer to that of the language results in erratic performances that do not scale well. Future work will include experiments to evaluate the effect of tokenizer post-retraining on the grammar token distribution for compact languages (e.g. TFL+) to understand the cases where re-training would help or fail.

\begin{table*}
  \caption{Supervised Fine-Tuning Results on L4 GPU: best results in bold and the second best underlined.}
  \label{tab:sftresultssmall}
    \resizebox{0.8\linewidth}{!}{
  \begin{tabular}{cccccc}
    \toprule
    Model&Grammar&Accuracy&Precision&Recall&F1\\
    \midrule
Flan-T5-small&\underline{\textbf{NL}}&\underline{0.4384}&\underline{0.4385}&\textbf{0.4355}&\textbf{0.4351}\\
Flan-T5-small&\underline{\textbf{CLIF}}&\underline{0.4384}&\textbf{0.4425}&\underline{0.4274}&\underline{0.4109}\\
Flan-T5-small&\underline{NL + FOL}&\underline{0.4384}&0.4078&0.4261&0.3951\\
Flan-T5-small&FOL&0.4236&0.3983&0.4108&0.3827\\
GPT-2$^*$&NL&0.3743&0.3675&0.3713&0.3677\\
GPT-2$^*$&NL + FOL&0.3793&0.3878&0.3844&0.3671\\
Flan-T5-small&CGIF&0.3743&0.3609&0.3666&0.3398\\
Flan-T5-large$^*$&\textbf{NL}&\textbf{0.4729}&0.3391&0.3396&0.3391\\
Flan-T5-small&TFL&0.3596&0.3424&0.3476&0.3070\\
Flan-T5-large$^*$&NL + FOL&0.4088&0.3566&0.2915&0.3017\\
Flan-T5-large$^*$&TFL&0.4334&0.2858&0.3112&0.2962\\
Flan-T5-small&TFL+&0.3596&0.3243&0.3440&0.2918\\
GPT-2$^*$&FOL&0.3349&0.3021&0.3260&0.2902\\
Flan-T5-large$^*$&TFL+&0.2758&0.2326&0.1988&0.2143\\
Flan-T5-large$^*$&FOL&0.3251&0.2096&0.2323&0.2086\\
Flan-T5-large$^*$&CLIF&0.2709&0.2297&0.1930&0.1949\\
Flan-T5-small&MINIFOL&0.3596&0.2008&0.26253&0.1904\\
Flan-T5-large$^*$&CGIF&0.3251&0.1828&0.2308&0.1811\\
Flan-T5-large$^*$&MINIFOL&0.2955&0.1895&0.2101&0.1771\\
  \bottomrule
\footnotesize{$^*$ The model is fine-tuned wth PEFT-LoRA}\\
\end{tabular}
}
\end{table*}

\begin{table*}
  \caption{Supervised Fine-Tuning Results on A100 GPU: best results in bold and second best underlined.}
  \label{tab:sftresultslarge}
  \begin{tabular}{cccccc}
    \toprule
    Model&Grammar&Accuracy&Precision&Recall&F1\\
    \midrule
Flan-T5-base&CLIF&0.5073&0.5001&0.5015&0.4986\\
Flan-T5-base&NL&0.5418&0.5425&0.5397&0.5387\\
Flan-T5-base&TFL&0.4827&0.6240&0.4646&0.4187\\
Flan-T5-large&\textbf{NL}&\textbf{0.6600}&\textbf{0.6622}&\textbf{0.6572}&\textbf{0.6585}\\
Flan-T5-large&\underline{CLIF}&\underline{0.6157}&\underline{0.6148}&\underline{0.6156}&\underline{0.6149}\\
Flan-T5-base&FOL&0.4876&0.4716&0.4727&0.4444\\
Flan-T5-base&TFL+&0.4926&0.5690&0.4775&0.4504\\
Flan-T5-large&TFL+&0.5418&0.5594&0.5350&0.5347\\
  \bottomrule
\end{tabular}
\end{table*}

\begin{table*}
  \caption{Zero-Shot Results on L4 GPU. The best results are in bold and the second best are underlined.}
  \label{tab:zsresults}
    \resizebox{0.75\linewidth}{!}{
  \begin{tabular}{cccccc}
    \toprule
    Model&Grammar&Accuracy&Precision&Recall&F1\\
    \midrule
Gemma-2-2b-it&\underline{\textbf{NL}}&\underline{0.4827}&0.4704&\underline{0.4901}&\textbf{0.4106}\\
Phi-3.5-mini-instruct&\underline{NL}&0.4380&0.5300&0.4200&\underline{0.4000}\\
Flan-T5-large&\textbf{NL}&\textbf{0.4975}&0.3313&\textbf{0.5026}&0.3991\\
Gemma-2-2b-it&CLIF&0.4433&0.5028&0.4346&0.3672\\
Phi-3.5-mini-instruct&NL + FOL&0.4090&0.5100&0.3900&0.3600\\
Gemma-2-2b-it&FOL&0.4334&0.4989&0.4233&0.3574\\
Gemma-2-2b-it&MINIFOL&0.3694&0.4081&0.3584&0.2937\\
Flan-T5-large&CLIF&0.3891&0.2899&0.3769&0.2828\\
Flan-T5-small&NL&0.3990&0.2959&0.3824&0.2669\\
GPT-2&FOL&0.3103&0.2993&0.3250&0.2660\\
Gemma-2-2b-it&CGIF&0.3399&0.3276&0.3260&0.2657\\
Phi-3.5-mini-instruct&FOL&0.3550&0.2400&0.3400&0.2500\\
GPT-2&NL&0.2758&0.2817&0.2881&0.2499\\
Gemma-2-2b-it&\textbf{TFL+}&0.3399&\textbf{0.5634}&0.3310&0.2498\\
GPT-2&\underline{NL + FOL}&0.3152&\underline{0.5444}&0.3335&0.2494\\
Flan-T5-large&CGIF&0.3399&0.2001&0.3261&0.2287\\
Flan-T5-large&NL + FOL&0.3596&0.1830&0.2635&0.2034\\
Gemma-2-2b-it&TFL&0.3399&0.2225&0.2419&0.1863\\
Flan-T5-small&FOL&0.3546&0.2294&0.3340&0.1842\\
Flan-T5-small&NL + FOL&0.3546&0.1182&0.3333&0.1745\\
Flan-T5-large&TFL&0.3546&0.1182&0.3333&0.1745\\
Flan-T5-large&MINIFOL&0.3497&0.0777&0.1012&0.0787\\
Flan-T5-large&FOL&0.3300&0.0658&0.1332&0.0779\\
Flan-T5-small&MINIFOL&0&0&0&0\\
Flan-T5-small&TFL&0&0&0&0\\
  \bottomrule
\end{tabular}
}
\end{table*}

\begin{table*}
  \caption{Zero-Shot with and without BNF Grammar Prompting Results on L4 GPU. The best results are in bold and the second best are underlined.}
  \label{tab:zsresultsgrammar}
  \begin{tabular}{ccccccc}
    \toprule
    Model&Grammar&Accuracy&Precision&Recall&F1&Grammar Prompting\\
    \midrule
Gemma-2-2b-it&\textbf{CLIF}&\textbf{0.4532}&0.4633&\textbf{0.4368}&\textbf{0.3924}&Yes\\
Gemma-2-2b-it&\underline{CLIF}&\underline{0.4433}&\underline{0.5028}&\underline{0.4346}&\underline{0.3672}&No\\
Gemma-2-2b-it&FOL&0.4187&0.4021&0.4018&0.3514&Yes\\
Gemma-2-2b-it&FOL&0.4334&0.4989&0.4233&0.3574&No\\
Gemma-2-2b-it&TFL+&0.3596&0.3618&0.3445&0.2782&Yes\\
Gemma-2-2b-it&\textbf{TFL+}&0.3399&\textbf{0.5634}&0.3310&0.2498&No\\
Gemma-2-2b-it&TFL&0.3645&0.4689&0.3458&0.2422&Yes\\
Gemma-2-2b-it&TFL&0.3399&0.2225&0.2419&0.1863&No\\
  \bottomrule
\end{tabular}
\end{table*}

\begin{table*}
  \caption{8-Shot Results on L4 GPU. The best results are in bold and the second best are underlined.}
  \label{tab:fsresults}
  \resizebox{0.7\linewidth}{!}{
  \begin{tabular}{cccccc}
    \toprule
    Model&Grammar&Accuracy&Precision&Recall&F1\\
    \midrule
Flan-T5-large&\textbf{NL}&\textbf{0.4926}&0.3313&\textbf{0.5002}&\textbf{0.3959}\\
Flan-T5-large&\underline{CLIF}&\underline{0.4334}&0.3302&\underline{0.4223}&\underline{0.3269}\\
Flan-T5-large&NL + FOL&0.3940&0.2773&0.3853&0.2989\\
Flan-T5-large&MINIFOL&0.3891&0.2485&0.3821&0.2963\\
Phi-3.5-mini-instruct&\textbf{NL}&0.2710&\textbf{0.5200}&0.2700&0.2700\\
Phi-3.5-mini-instruct&\underline{FOL}&0.2320&\underline{0.3700}&0.2300&0.2600\\
Flan-T5-small&NL&0.3940&0.3674&0.3763&0.2548\\
Flan-T5-large&FOL&0.3546&0.2107&0.3355&0.2016\\
Flan-T5-large&TFL&0.3251&0.2045&0.3085&0.1934\\
Flan-T5-small&FOL&0.3546&0.1182&0.3333&0.1745\\
Flan-T5-small&NL + FOL&0.3546&0.1182&0.3333&0.1745\\
GPT-2&FOL&0.3546&0.1182&0.3333&0.1745\\
Gemma-2-2b-it&NL&0.3546&0.1182&0.3333&0.1745\\
Gemma-2-2b-it&TFL&0.3546&0.1182&0.3333&0.1745\\
Gemma-2-2b-it&TFL+&0.3546&0.1182&0.3333&0.1745\\
Gemma-2-2b-it&FOL&0.3546&0.1182&0.3333&0.1745\\
Gemma-2-2b-it&CLIF&0.3546&0.1182&0.3333&0.1745\\
Gemma-2-2b-it&CGIF&0.3546&0.1182&0.3333&0.1745\\
GPT-2&NL&0.3399&0.1133&0.3333&0.1691\\
GPT-2&NL + FOL&0.3054&0.1018&0.3333&0.1559\\
Flan-T5-large&CGIF&0.3448&0.1467&0.2447&0.1488\\
Phi-3.5-mini-instruct&NL + FOL&0.1380&0.1200&0.1400&0.1300\\
Flan-T5-small&MINIFOL&0.1231&0.0054&0.0041&0.0047\\
Flan-T5-small&TFL&0&0&0&0\\
  \bottomrule
\end{tabular}
}
\end{table*}

\begin{table*}
  \caption{8-Shot with and without BNF Grammar Prompting Results on L4 GPU.}
  \label{tab:fsresultsgrammar}
  \resizebox{0.9\linewidth}{!}{
  \begin{tabular}{ccccccc}
    \toprule
    Model&Grammar&Accuracy&Precision&Recall&F1&Grammar Prompting\\
    \midrule
Gemma-2-2b-it&TFL&0.3546&0.1182&0.3333&0.1745&Yes\\
Gemma-2-2b-it&TFL&0.3546&0.1182&0.3333&0.1745&No\\
Gemma-2-2b-it&TFL+&0.3546&0.1182&0.3333&0.1745&Yes\\
Gemma-2-2b-it&TFL+&0.3546&0.1182&0.3333&0.1745&No\\
Gemma-2-2b-it&FOL&0.3546&0.1182&0.3333&0.1745&Yes\\
Gemma-2-2b-it&FOL&0.3546&0.1182&0.3333&0.1745&No\\
Gemma-2-2b-it&CLIF&0.3546&0.1182&0.3333&0.1745&Yes\\
Gemma-2-2b-it&CLIF&0.3546&0.1182&0.3333&0.1745&No\\
  \bottomrule
\end{tabular}
}
\end{table*}

\begin{table*}
  \caption{Supervised Fine-Tuning with and without Grammar Context-Passing Results. The best results are in bold and the second best are underlined.}
  \label{tab:sftresultsgrammar}
  \resizebox{0.9\linewidth}{!}{
  \begin{tabular}{llllllll}
    \toprule
    Model&Grammar&Accuracy&Precision&Recall&F1&ICGP$^\dagger$&GPU\\
    \midrule
Flan-T5-large$^*$&TFL&0.3546&0.1182&0.3333&0.1745&Yes&L4\\
Flan-T5-large$^*$&TFL&0.4334&0.2858&0.3112&0.2962&No&L4\\
Flan-T5-small&CLIF&0.3694&0.3131&0.3486&0.2299&Yes&L4\\
Flan-T5-small&\underline{CLIF}&\underline{0.4384}&0.4425&\underline{0.4274}&\underline{0.4109}&No&L4\\
Flan-T5-base&\textbf{CLIF}&0.4334&\textbf{0.6252}&0.4129&0.3317&Yes&A100\\
Flan-T5-base&\underline{\textbf{CLIF}}&\textbf{0.5073}&\underline{0.5001}&\textbf{0.5015}&\textbf{0.4986}&No&A100\\
  \bottomrule
  \footnotesize{$^*$ The model is fine-tuned wth PEFT-LoRA}\\
  \footnotesize{$^\dagger$ ICGP = In-Context Grammar Passing}\\
\end{tabular}
}
\end{table*}

\begin{table*}
  \caption{Supervised Fine-Tuning with Tokenizer Re-Training Results. The best results are in bold and the second best are underlined.}
  \label{tab:sftresultsvocab}
  \resizebox{\linewidth}{!}{
  \begin{tabular}{ccccccccc}
    \toprule
    Model&Grammar&Accuracy&Precision&Recall&F1&GPU&Re-Train Tokenizer&Vocabulary Size\\
    \midrule
Flan-T5-small&TFL&0.3201&0.2983&0.3109&0.2790&T4/L4&Yes&191\\
Flan-T5-small&TFL&0.3596&0.3424&0.3476&0.3070&T4/L4&No&32128\\
Flan-T5-small&CLIF&0.3497&0.3600&0.3466&0.3382&T4/L4&Yes&32128\\
Flan-T5-small&CLIF&0.4384&0.4425&0.4274&0.4109&T4/L4&No&32128\\
Flan-T5-small&TFL+&0.4532&0.4286&0.4395&0.4113&T4/L4&Yes&180\\
Flan-T5-small&TFL+&0.3596&0.3243&0.3440&0.2918&T4/L4&No&32128\\
Flan-T5-base&TFL+&0.4334&0.4083&0.4168&0.3713&A100&Yes&180\\
Flan-T5-base&\underline{\textbf{TFL+}}&\underline{0.4926}&\textbf{0.5690}&\underline{0.4775}&\underline{0.4504}&A100&No&32128\\
Flan-T5-large&TFL+&0.4039&0.4786&0.3854&0.3167&A100&Yes&180\\
Flan-T5-large&\underline{\textbf{TFL+}}&\textbf{0.5418}&\underline{0.5594}&\textbf{0.5350}&\textbf{0.5347}&A100&No&32128\\
  \bottomrule
\end{tabular}
}
\end{table*}

\begin{table*}
  \caption{Syllogism Evaluation Framework for Supervised Fine-Tuning Flan-T5-large on NL, CLIF and TFL+.}
\label{tab:syllogism-evaluation}
  \begin{tabular}{|c|c|c|c|c|c|c|c|c|}
    \hline
    \multirow{2}{*}{Grammar} &
      \multicolumn{2}{c}{Hypothetical} &
      \multicolumn{2}{c}{Disjunctive} &
      \multicolumn{2}{c}{Complex} &
      \multicolumn{2}{c|}{Categorical} \\
    & Hit & Miss & Hit & Miss & Hit & Miss & Hit & Miss \\
    \hline
    NL & 59 & 25 & 60 & 39 & 14 & 2 & 2 & 2 \\
    \hline
    CLIF & 48 & 36 & 57 & 42 & 14 & 2 & 2 & 2 \\
    \hline
    TFL+ & 44 & 40 & 53 & 46 & 15 & 1 & 2 & 2 \\
    \hline
  \end{tabular}
\end{table*}

\begin{table*}
  \caption{Flan-T5-large model SFT performance on NL, CLIF and TFL+ for Categorical syllogisms.}
  \label{tab:syllogismerroranalysis}
\resizebox{\linewidth}{!}{
  \begin{tabular}{p{0.4\linewidth} | p{0.2\linewidth}|c|c|c|c}
    \toprule
    Premises&Conclusion&True Label&NL Label&CLIF Label&TFL+ Label\\
    \midrule
Some affection is love. Some love is positive.&Some affection is positive.&Uncertain&True&True&True\\
\hline
Heinrich Schmidt was a German politician. 
Heinrich Schmidt was also a member of the Prussian State Parliament and the Nazi Reichstag.&Heinrich Schmidt was German or Russian or both.&True&Uncertain&Uncertain&False\\
\hline
Heinrich Schmidt was a German politician. 
Heinrich Schmidt was also a member of the Prussian State Parliament and the Nazi Reichstag.&Some German politician was part of both the Prussian State Parliament and the Nazi Reichstag.&True&True&True&True\\
\hline
Heinrich Schmidt was a German politician. 
Heinrich Schmidt was also a member of the Prussian State Parliament and the Nazi Reichstag.&No politicians are part of the Nazi Reichstag.&False&False&False&False\\
  \bottomrule
\end{tabular}
}
\end{table*}

\section{Conclusions and Future Work}
\label{section-conclusion}

In this paper, we presented the early stage of a PhD research tackling the formal representation of data in SLMs to possibly bootstrap the construction of ontologies with external knowledge. We focused here on the formalization of logical data with the goal of finding a compact representation that keeps SLMs performing competitively on reasoning tasks. Our experiments enable us to answer our RQ: logical data for first-order reasoning in NL can be challenged by more compact and formal representations. From our empirical results, CLIF emerged as a strong candidate due to the compactness of its vocabulary, its easy transformation to and from FOL and its competitive results especially in a SFT setting. The results were achieved on small, frugal language models that are all under 3 billion parameters, making them accessible for reasoning tasks over logical data. While our experiments do not point to an absolute best representation that overtakes NL in the formalization of logical data, they show that expressing inputs in compact formal languages comes very close for SLMs on reasoning tasks. This takeaway allows us to focus on the next goal of this PhD research which is to leverage SLMs and formal representations of ontological data to bootstrap ontology construction. 

\subsection{Next Activities in this PhD Research}
\label{subsection-nextactivities}
This PhD research will explore mixed-input representations (e.g. NL + CLIF) to assess whether combining expressiveness with the structure of a formal grammar improves reasoning. We will also explore injecting knowledge from a high-level ontology (e.g. DOLCE \cite{borgo2022dolce}) in a formal language to efficiently reuse relevant knowledge learned by the model in an ontology extension pipeline.

\subsection{Related Future Work Directions}
\label{subsection-futuredirections}

We identify several future research directions including the extension of the SEF-CLGC evaluation to datasets such as ProofWriter\footnote{https://www.kaggle.com/datasets/mathurinache/proofwriter}, RuleTaker\footnote{https://github.com/allenai/ruletaker}, the Logical Entailment Dataset{\footnote{https://github.com/google-deepmind/logical-entailment-dataset} and SynLogic\footnote{https://github.com/MiniMax-AI/SynLogic}. Another direction would be using LMs to extend the vocabulary of an ontology to handle new scenarios given Competency Questions (CQs) and the original ontology in practical use cases like HMAS\footnote{https://github.com/HyperAgents/hmas} using categorical syllogisms to explain the hierarchical connections between generated classes and properties.

\begin{acknowledgments}
This work is supported by 3IA Côte d’Azur (ANR-19-P3IA-0002), UCAJEDI (ANR-15-IDEX-01), Université Côte d’Azur’s Center for High-Performance Computing and Data ScienceTech Institute.
\end{acknowledgments}

\section*{Declaration on Generative AI}
  
The author(s) have not employed any Generative AI tools.
  
\bibliography{sample-ceur}

\appendix

\end{document}